\documentclass[11pt]{article}
\usepackage{acl}
\usepackage{times}
\usepackage{latexsym}
\usepackage{booktabs}
\usepackage[T1]{fontenc}
\usepackage[utf8]{inputenc}
\usepackage{microtype}
\usepackage{inconsolata}
\usepackage{graphicx}
\usepackage{amsmath}
\usepackage{amssymb}
\usepackage{algorithm}
\usepackage{algorithmic}
\usepackage{listings}
\usepackage{xcolor}
\usepackage{multirow}
\usepackage{url}

\definecolor{BlueBG}{rgb}{0,0.46,0.71}

\title{GraphBit: A Graph-based Agentic Framework for Non-Linear Agent Orchestration}

\author{
Yeahia Sarker$^{1}$
Md Rahmat Ullah$^{2}$
Musa Molla$^{2}$
Shafiq Joty$^{3}$ \\
$^{1}$MTSU \\
$^{2}$InfinitiBit GmbH \\
$^{3}$Salesforce Research \\
\small\texttt{
ys5d@mtmail.mtsu.edu, 
{rahmat.ullah@infinitibit.com, musa.molla}@infinitibit.com \and
srjoty@ntu.edu.sg
}
}

\begin{document}

\maketitle

\begin{abstract}

Agentic LLM frameworks that rely on prompted orchestration—where the model itself determines workflow transitions—often suffer from hallucinated routing, infinite loops, and non-reproducible execution. We introduce GraphBit, an engine-orchestrated framework that defines workflows explicitly and deterministically as a directed acyclic graph (DAG). Unlike prompted orchestration, agents in GraphBit operate as typed functions, while a Rust-based engine governs routing, state transitions, and tool invocation, ensuring reproducibility and auditability. The engine supports parallel branch execution, conditional control flow over structured state predicates, and configurable error recovery. A three-tier memory architecture—ephemeral scratch space, structured state, and external connectors—isolates context across stages, preventing cascading context bloat that degrades reasoning in long-running pipelines. Across GAIA benchmark tasks spanning zero-tool, document-augmented, and web-enabled workflows, GraphBit outperforms six existing frameworks, achieving the highest accuracy (67.6\%), zero framework-induced hallucinations, the lowest latency (11.9 ms overhead), and the highest throughput. Ablation studies demonstrate that each memory tier contributes measurably to performance, with deterministic execution providing the greatest gains on tool-intensive tasks representative of real-world deployments.

{\href{}{\texttt{Code: github.com/InfinitiBit/graphbit}}
}


\end{abstract}

\section{Introduction}

The emergence of Large Language Model (LLM)-based agents marks a paradigm shift in AI, combining foundation model reasoning capabilities with environmental perception, decision-making, and autonomous action execution \cite{yao2022react,ke2025surveyfrontiersllmreasoning}. Multi-agent systems have rapidly evolved from research prototypes to production deployments, with applications spanning software engineering \cite{hong2023metagpt}, scientific discovery \cite{boiko2023autonomous}, and enterprise automation \cite{wu2024autogen}. These systems decompose complex tasks into specialized subtasks assigned to collaborative agents that coordinate to achieve collective goals \cite{guo2024large}. A central challenge in operationalizing these systems is \textit{workflow orchestration}-- specifying which agents to invoke, in what order, which tools to employ, and how data propagates across stages. The underlying framework must execute this workflow reliably, efficiently, and reproducibly \cite{gu2025large,tran2025multi}.

\begin{figure}[t!]
    \centering
    \includegraphics[width=0.98\linewidth]{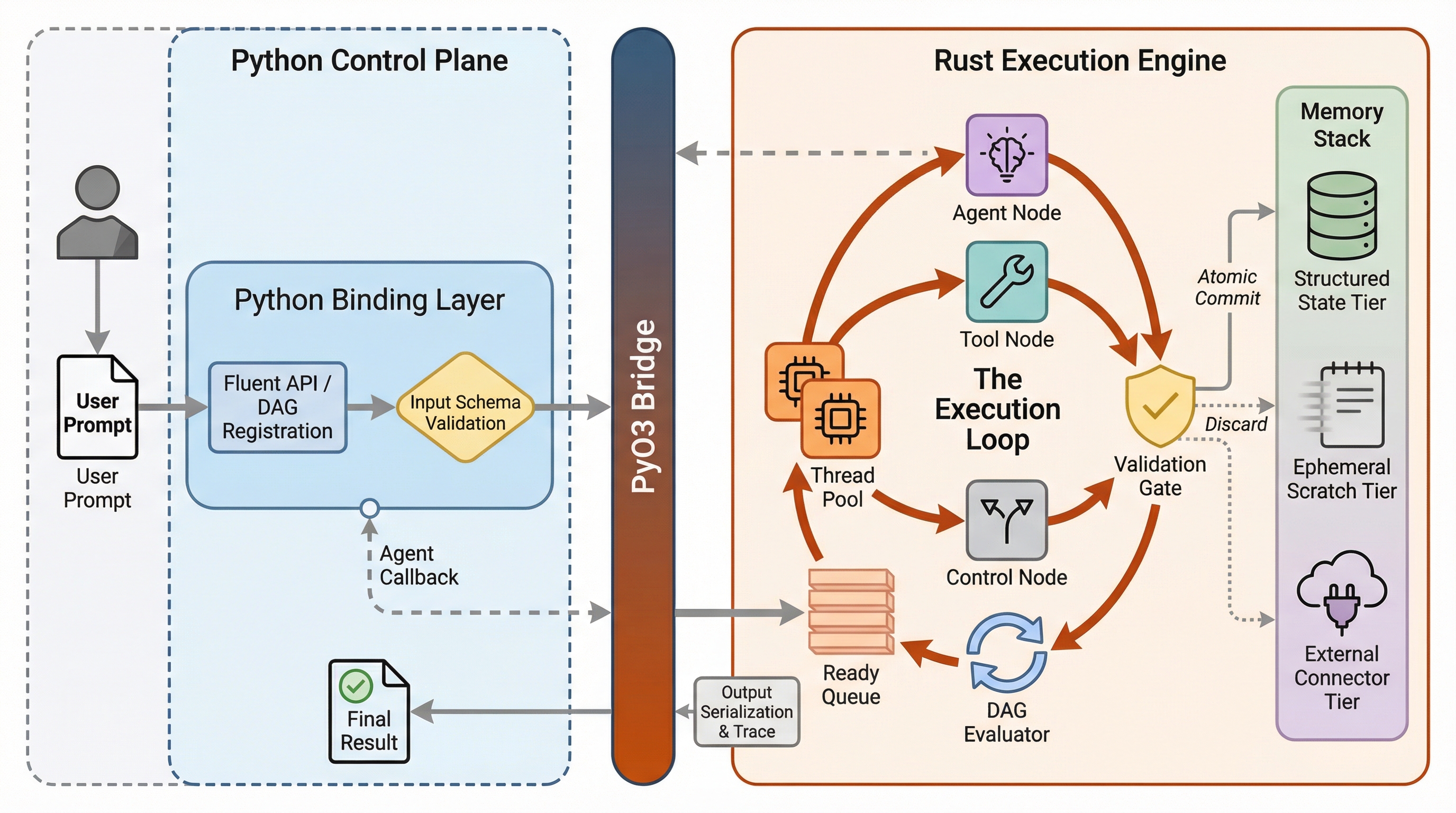}
    \caption{Overview of GraphBit's engine-based orchestration. Given a user-defined workflow graph, the execution engine deterministically governs all routing and state transitions, while agents focus solely on domain-specific reasoning within their assigned nodes.}
    \label{fig:visual_comparison}
\end{figure}

Despite their promise, most existing multi-agent frameworks suffer from a fundamental architectural limitation: they rely on \emph{prompted-orchestration} where the LLM itself determines workflow transitions. This includes frameworks such as LangChain \cite{annam2025langchain}, CrewAI \cite{duan2024exploration}, and AutoGen \cite{wu2024autogen}, where agents receive natural language descriptions of available tools and downstream agents, then select their next action through in-context learning. This design introduces three critical failure modes \cite{cemri2025multi,patil2024gorilla}: (1)~hallucinated routing, where the LLM invents non-existent agents or tools, causing silent failures; (2)~infinite loops, where agents repeatedly invoke each other without architectural termination conditions; and (3)~non-deterministic execution, where identical inputs produce different traces, undermining auditability in regulated domains. Additionally, each orchestration decision requires a full LLM inference pass, and memory scales with accumulated context, creating inefficiencies that often become acute in enterprise settings with strict latency budgets \cite{sculley2015hidden,barua2024exploring}.
\par
We present GraphBit, an \emph{engine-orchestrated} agentic framework that performs multi-agent orchestration through a graph-based, non-linear execution paradigm. In GraphBit, users define workflows as typed directed acyclic graphs (DAGs) specifying agent nodes, tool nodes, and control-flow logic. The framework then executes this workflow deterministically: agents operate as typed functions responsible solely for their domain-specific reasoning, while the execution engine governs all workflow transitions, state management, and tool invocations according to the user-specified graph. This separation ensures that given any workflow, execution remains predictable, auditable, and reproducible regardless of the stochastic nature of LLM outputs \cite{qiu2025blueprint}.
\par
GraphBit's architecture rests on three foundational principles: (1)~\textit{graph-native execution}, where workflows are expressed as DAGs with typed edges representing data dependencies and control flow, enabling parallel execution of independent branches; (2)~\textit{engine-governed orchestration}, where the execution engine makes all routing decisions based on explicit conditions, eliminating hallucinated routing and infinite loops by construction; and (3)~\textit{hierarchical memory isolation}, where a three-tier memory model segregates ephemeral scratch space, structured workflow state, and external connector interfaces, preventing context pollution. This has been illustrated in Figure~\ref{fig:visual_comparison}. Our Rust-based execution core with Python bindings achieves 11.9\,ms mean processing latency and 5,025 operations per minute throughput, a 3$\times$ improvement over the fastest comparable baseline.
\par
The contributions of this paper are fourfold: (1) an engine-based orchestration architecture with a three-tier memory model where a deterministic execution engine governs all transitions within a user-defined workflow graph, eliminating hallucinated routing by construction; (2)~a comprehensive evaluation of seven LLM agent frameworks on a curated 68-task GAIA benchmark subset \cite{mialon2023gaia} spanning three workflow types, where all frameworks execute equivalent workflow configurations; (3)~demonstration that given equivalent workflows, GraphBit attains 67.6\% accuracy with 0\% hallucination rate, outperforming the strongest baseline by 14.7 percentage points; and (4)~ablation studies isolating the contribution of each architectural component to overall performance.

\section{Related Work}

We situate GraphBit within the landscape of agent architectures, multi-agent frameworks, and workflow orchestration systems.

\textbf{Agent Architectures.} The ReAct paradigm \cite{yao2022react} established the foundation for modern LLM agents by interleaving reasoning traces with action execution, building on chain-of-thought prompting \cite{wei2022chain,masterman2024landscape}. Subsequent work extended this through Tree of Thoughts \cite{yao2023tree,ranaldi2024tree} for parallel reasoning paths and Reflexion \cite{shinn2023reflexion} for self-reflective improvement. Toolformer \cite{schick2023toolformer} demonstrated that LLMs can learn tool invocation through fine-tuning, while later work showed in-context learning suffices for capable models \cite{qin2024tool,qin2023toolllm}. GraphBit builds on these insights by treating agents as specialized tool-invoking functions, but crucially separates tool selection (performed by the agent) from workflow orchestration (performed by the engine). 

\textbf{Multi-Agent Frameworks.} MetaGPT \cite{hong2023metagpt} coordinates agents through standardized operating procedures for software engineering roles. ChatDev \cite{qian2024chatdev} organizes agents into a virtual software company with defined communication protocols. AutoGen \cite{wu2024autogen} provides a conversation-centric framework with natural language agent interaction. LangChain \cite{annam2025langchain} and its graph-oriented extension LangGraph \cite{wang2024agent} represent the most widely adopted orchestration frameworks; LangGraph introduces explicit graph structures but retains LLM-based routing at conditional edges. Recent work on parallel function calling \cite{kim2024llm} formulates tool dispatch as a DAG but does not address multi-agent orchestration. LlamaIndex \cite{Liu_LlamaIndex_2022} focuses on retrieval-augmented generation pipelines, while Pydantic AI\footnote{\url{https://ai.pydantic.dev/}} provides type-safe agent definitions with structured output validation. All these frameworks share a common assumption: the LLM participates in orchestration decisions, which recent empirical analysis has shown leads to systematic failure modes including task verification gaps and inter-agent misalignment \cite{cemri2025multi}. GraphBit departs from this paradigm by restricting the LLM to domain-specific reasoning, delegating all orchestration to a deterministic execution engine. Yu \textit{et al.} \cite{yu2025dyntaskmas} introduce DynTaskMAS, a dynamic task-graph framework for LLM-based multi-agent systems that supports adaptive decomposition, parallel execution, context sharing, and workflow optimization. It improves scalability and efficiency in complex, evolving tasks. However, increased orchestration and synchronization overhead may reduce practicality in simpler or latency-sensitive settings.
\par
\textbf{Workflow Orchestration.} Traditional workflow engines such as Apache Airflow \cite{haines2022workflow} and Prefect \cite{narayanan2024orchestrating} provide deterministic execution but lack native LLM agent support. Temporal\footnote{\url{https://temporal.io/}} offers durable execution with automatic retry and state persistence but requires substantial integration effort. DSPy \cite{khattab2023dspy} compiles declarative language programs into optimized prompts, and LMQL \cite{beurer2023prompting} introduces constrained LLM generation. These approaches improve individual agent reliability but do not address multi-agent orchestration. GraphBit complements these techniques by providing a reliable orchestration layer that can incorporate any agent implementation.

\section{System Architecture}

\begin{figure*}[t!]
    \centering
    \includegraphics[width=0.92\linewidth]{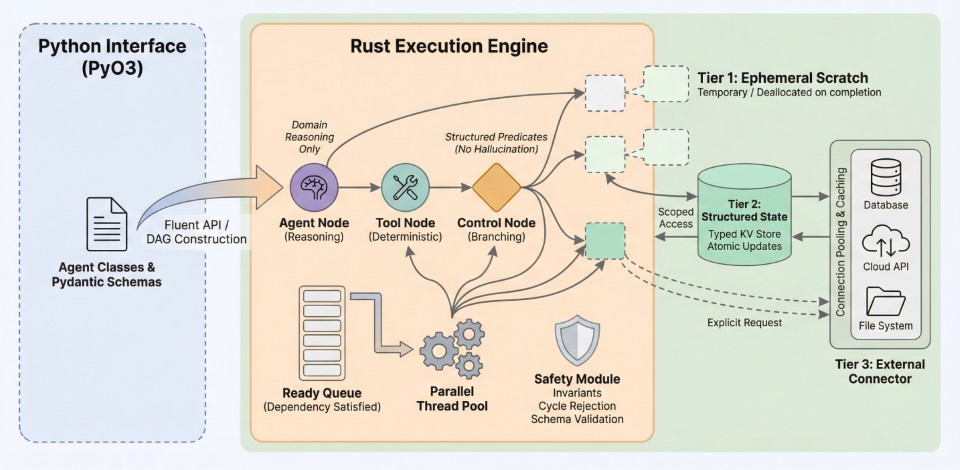}
    \caption{Detailed GraphBit architecture. The Rust-based execution engine traverses a typed workflow DAG, dispatching agent, tool, and control nodes while the three-tier memory system (ephemeral scratch, structured state, external connectors) isolates context across execution stages.}
    \label{fig:architecture}
\end{figure*}

GraphBit comprises four integrated components: a workflow graph specification, a Rust-based execution engine, a three-tier memory system, and Python bindings for agent development (Figure~\ref{fig:architecture}).

\subsection{Workflow Graph Specification}

Workflows in GraphBit are expressed as directed acyclic graphs (DAGs) where nodes represent computational units and typed edges encode data dependencies and control flow \cite{qiao2024benchmarking}. Three node types compose to express complex multi-agent workflows \cite{thost2021directed}. \textit{Agent nodes} encapsulate LLM-based reasoning units, each specifying input/output schemas, a system prompt, and an optional tool set. The execution engine invokes the underlying LLM only when all input dependencies are satisfied, with type enforcement at node boundaries preventing schema violations from propagating through the workflow. \textit{Tool nodes} represent deterministic functions (web search, database queries, API calls) that execute without LLM inference, providing predictable latency. Tool nodes may be composed into subgraphs encapsulating complex retrieval or transformation logic. \textit{Control nodes} implement workflow logic including conditional branching, parallel fan-out, and aggregation. Crucially, control node decisions are evaluated by the execution engine against structured state predicates, not by LLM inference. For example, a conditional branch evaluates a boolean expression over workflow state variables rather than prompting an LLM to select the next step. Edges carry typed data between nodes with automatic serialization for cross-language interoperability, and optional transformation functions enable lightweight preprocessing during data transfer.

\subsection{Execution Engine}

The execution engine implements a dataflow model \cite{dennis2005first} optimized for LLM workloads. Written in Rust for performance and memory safety, the engine maintains a ready queue of nodes whose input dependencies are satisfied, dispatching independent nodes in parallel across a thread pool while executing dependent nodes sequentially with automatic data transfer between stages \cite{bugden2022rust}. The engine enforces correctness invariants that eliminate common failure modes in prompt-orchestrated systems. Termination guarantees arise from the DAG structure: cycles are rejected at graph construction time, and execution progress is tracked to detect stalls. Deterministic routing follows from evaluating control predicates against structured state rather than LLM outputs. Type safety is enforced at node boundaries through runtime schema validation \cite{polyzotis2019data,habib2019type}, with violations raised as explicit errors. Error handling follows a fail-fast philosophy with configurable recovery policies ranging from immediate termination to automatic retry with exponential backoff, and checkpointing enables resumption from intermediate states for long-running workflows.

\subsection{Three-Tier Memory Architecture}
GraphBit's memory system comprises three isolated tiers designed to prevent context pollution while enabling efficient data access \cite{zhang2025survey}. The \textit{ephemeral scratch tier} provides temporary storage for intermediate computations within a single node execution, allocated at node start and deallocated upon completion. This prevents implementation details from leaking across node boundaries, particularly for agents performing iterative chain-of-thought reasoning. The \textit{structured state tier} maintains the canonical workflow context as a typed key-value store with atomic updates upon successful node completion. The engine tracks state provenance for full auditability, and scoped access ensures nodes may only read explicitly declared state keys, preventing implicit dependencies. The \textit{external connector tier} provides managed interfaces to databases, APIs, and file systems with connection pooling, automatic retry, and result caching. Connector results are not automatically injected into agent contexts; nodes must explicitly request external data, preventing context bloat. This three-tier architecture prevents cascading context growth that degrades LLM performance, enables reproducible execution through explicit state management, and simplifies testing through connector abstraction.

\vspace{-0.3em}
\subsection{Python Bindings and Agent Development}
Agent development occurs in Python through PyO3 bindings, combining Rust's performance for orchestration with Python's ecosystem for LLM integration. Developers define agent classes specifying input/output schemas as Pydantic models and implement a \texttt{run} method containing domain-specific logic; the framework handles serialization, context assembly, and LLM invocation. Tool integration follows a similar declarative pattern with support for asynchronous execution and streaming. Workflow composition uses a fluent API that constructs the underlying DAG, with graph structure and type compatibility validated at construction time. Complex workflows may be encapsulated as reusable subgraphs for modular design.

\vspace{-0.3em}
\section{Experimental Evaluation}
\vspace{-0.5em}
We evaluate GraphBit against six widely-adopted LLM agent frameworks: LangChain \cite{annam2025langchain}, LangGraph \cite{annam2025langchain}, CrewAI \cite{duan2024exploration}, Microsoft AutoGen \cite{wu2024autogen}, Pydantic AI, and LlamaIndex \cite{Liu_LlamaIndex_2022}. Our evaluation addresses four research questions: (RQ1)~How does GraphBit compare to existing frameworks on task completion accuracy across diverse task types? (RQ2)~What are the computational efficiency gains from Rust-based orchestration? (RQ3)~Does deterministic orchestration eliminate workflow reliability failures? (RQ4)~How does each architectural component contribute to overall performance?

\subsection{Experimental Setup}

We evaluate on the GAIA benchmark \cite{mialon2023gaia}, a comprehensive evaluation suite for general AI assistants on real-world tasks requiring multi-step reasoning, tool use, and web navigation. From the original 165 tasks, we curate a high-quality subset of 68 tasks by excluding tasks on which all seven frameworks consistently failed during preliminary testing, yielding a discriminative evaluation set. The curated set spans three difficulty levels: Level~1 contains 29 simple single-step tasks, Level~2 contains 36 moderate multi-step reasoning tasks, and Level~3 contains 3 complex planning tasks requiring extensive tool use.
\par
Critically, we segment the evaluation into three distinct workflow types that reflect real-world deployment patterns: (1)~\textbf{zero-tool tasks} (7~tasks) requiring pure LLM reasoning without external tool invocation; (2)~\textbf{document-augmented tasks} (19~tasks) requiring local tool invocation to process attached files (PDFs, spreadsheets, images); and (3)~\textbf{web-enabled tasks} (42~tasks) utilizing web search for real-time information retrieval. This segmentation enables fine-grained analysis of framework capabilities across fundamentally different agentic scenarios. All frameworks use GPT-5.2 as the underlying LLM with identical temperature and sampling parameters. Task correctness is evaluated through dual verification: exact string matching against ground truth and independent LLM-based evaluation. We report six metrics: accuracy, hallucination rate (percentage of failed executions due to framework errors), mean processing time (framework overhead excluding LLM API latency), CPU utilization, peak memory consumption, and throughput (operations per minute). The complete experimental setup is provided in Appendix~\ref{sec:appendix_benchmark}.

\vspace{-0.4em}
\subsection{Overall Performance Comparison}

Table~\ref{tab:gaia_benchmark} presents the comparative results. GraphBit achieves the highest overall accuracy at 67.6\%, a 14.7 percentage point improvement over the strongest baseline (Pydantic~AI at 52.9\%). GraphBit is the only graph-based framework to achieve a 0\% hallucination rate, shared only with Pydantic~AI and LlamaIndex which employ simpler non-routing architectures. Processing latency of 11.9\,ms is 1.3$\times$ faster than LlamaIndex (15.0\,ms) and 5.9$\times$ faster than AutoGen (70.0\,ms), with throughput of 5,025 ops/min. Memory consumption of 126.1\,MB is 24\% lower than the closest baseline. These gains stem from the Rust execution engine, which eliminates Python interpreter overhead during orchestration.

\begin{table}[t!]
\centering
\small
\resizebox{\columnwidth}{!}{%
\begin{tabular}{@{}lccccc@{}}
\toprule
\textbf{Framework} & \textbf{Acc.} & \textbf{Hall.} & \textbf{Proc.} & \textbf{CPU} & \textbf{Mem.} \\
 & (\%) & (\%) & (ms) & (\%) & (MB) \\
\midrule
LangChain & 38.2 & 41.2 & 36.1 & 24.7 & 234.4 \\
LangGraph & 36.8 & 47.1 & 31.5 & 26.1 & 208.0 \\
CrewAI & 44.9 & 14.3 & 31.0 & 30.7 & 202.2 \\
AutoGen & 35.3 & 33.8 & 70.0 & 27.0 & 274.8 \\
Pydantic AI & 52.9 & 0.0 & 18.3 & 24.2 & 166.5 \\
LlamaIndex & 50.0 & 0.0 & 15.0 & 25.2 & 165.4 \\
\midrule
\textbf{GraphBit} & \textbf{67.6} & \textbf{0.0} & \textbf{11.9} & \textbf{21.1} & \textbf{126.1} \\
\bottomrule
\end{tabular}}
\caption{Overall performance on 68 curated GAIA tasks. Acc.: task completion accuracy; Hall.: hallucination rate (framework-induced failures); Proc.: mean processing time (framework overhead); CPU: average CPU utilization; Mem.: peak memory usage. Best results in \textbf{bold}.}
\label{tab:gaia_benchmark}
\vspace{-1.5em}
\end{table}

\vspace{-0.6em}
\subsection{Performance by Task Type}
Table~\ref{tab:task_type} disaggregates accuracy and hallucination by workflow type, revealing three key findings. First, on zero-tool tasks, four frameworks achieve identical 57.1\% accuracy with 0\% hallucination, indicating that orchestration differences are less impactful without tool routing. Second, GraphBit ties with LlamaIndex at 68.4\% on document-augmented tasks. Third, GraphBit's advantage is most pronounced on web-enabled tasks (69.0\% vs.\ 54.8\% for Pydantic~AI), which constitute 61.8\% of the evaluation set. The hallucination analysis reveals that LangGraph exhibits 69.0\% hallucination on web-enabled tasks, meaning over two-thirds of executions fail due to framework-induced errors. GraphBit's engine-governed tool invocation eliminates this failure mode entirely.

\begin{table}[t!]
\centering
\small
\resizebox{\columnwidth}{!}{%
\begin{tabular}{@{}lcccccc@{}}
\toprule
 & \multicolumn{2}{c}{\textbf{No-Tool}} & \multicolumn{2}{c}{\textbf{Local}} & \multicolumn{2}{c}{\textbf{Web}} \\
\cmidrule(lr){2-3} \cmidrule(lr){4-5} \cmidrule(lr){6-7}
\textbf{Framework} & Acc. & Hal. & Acc. & Hal. & Acc. & Hal. \\
\midrule
LangChain & 57.1 & 0.0 & 57.9 & 15.8 & 26.2 & 59.5 \\
LangGraph & 57.1 & 0.0 & 63.2 & 15.8 & 21.4 & 69.0 \\
CrewAI & 28.6 & 0.0 & 52.6 & 0.0 & 47.6 & 16.7 \\
AutoGen & 28.6 & 0.0 & 42.1 & 42.1 & 33.3 & 35.7 \\
Pydantic AI & 28.6 & 0.0 & 57.9 & 0.0 & 54.8 & 0.0 \\
LlamaIndex & 57.1 & 0.0 & 68.4 & 0.0 & 40.5 & 0.0 \\
\midrule
\textbf{GraphBit} & \textbf{57.1} & \textbf{0.0} & \textbf{68.4} & \textbf{0.0} & \textbf{69.0} & \textbf{0.0} \\
\bottomrule
\end{tabular}}
\caption{Accuracy (Acc.) and hallucination rate (Hal.) by task type (\%). No-Tool: 7 pure reasoning tasks; Local: 19 document-augmented tasks; Web: 42 web-search tasks. Best results in \textbf{bold}.}
\label{tab:task_type}
\vspace{-2.0em}
\end{table}

\vspace{-0.3em}
\subsection{Performance by Difficulty Level}

GraphBit achieves the highest accuracy on Level~1 (79.3\%) and Level~2 (63.9\%) tasks, outperforming AutoGen by 20.7 points on Level~1. Prompt-orchestrated frameworks degrade sharply with complexity: LangGraph drops from 48.3\% (Level~1) to 27.8\% (Level~2), and both LangGraph and AutoGen reach 0\% on Level~3. Pearson correlation analysis confirms statistically significant negative correlations for LangGraph ($r{=}{-}0.26$, $p{=}0.032$) and AutoGen ($r{=}{-}0.27$, $p{=}0.028$), while GraphBit shows no significant degradation ($p{>}0.05$). Full results by difficulty level are in Appendix~\ref{sec:appendix_difficulty}.

\subsection{Workflow Reliability Analysis}

We define hallucination as any framework-induced execution failure, including routing to non-existent agents, infinite loops, tool invocation failures, and unrecoverable runtime errors. The hallucination rates in Table~\ref{tab:task_type} reveal three key findings. First, no framework hallucinates on zero-tool tasks, confirming that pure LLM inference without tool routing is inherently stable. Second, hallucination rates escalate dramatically with tool complexity: LangGraph rises from 0\% (no-tool) to 15.8\% (local) to 69.0\% (web). Third, while GraphBit, Pydantic~AI, and LlamaIndex all achieve 0\% hallucination, only GraphBit combines this reliability with the highest accuracy (67.6\%), demonstrating that deterministic orchestration provides reliability without sacrificing reasoning quality. GraphBit eliminates hallucination by construction: the execution engine governs all state transitions according to the declared workflow graph, making it architecturally impossible for the LLM to route to non-existent agents or create execution cycles. The complete hallucination breakdown is provided in Appendix~\ref{sec:appendix_reliability}.

\subsection{Computational Efficiency Analysis}

GraphBit achieves the lowest processing latency across all three task types: 6.0\,ms on zero-tool tasks (yielding 10,000 ops/min), 10.8\,ms on document-augmented tasks, and 13.4\,ms on web-enabled tasks. Baselines exhibit steeper scaling; for instance, AutoGen reaches 159.1\,ms on document-augmented tasks due to conversation-based orchestration requiring multiple LLM round-trips. Memory scales efficiently from 34.9\,MB (no-tool) to 150.5\,MB (web), compared to AutoGen's 171.3--359.7\,MB range. Per-task-type efficiency breakdowns are provided in Appendix~\ref{sec:appendix_results}.

\subsection{Memory Architecture Ablation}

We validate our architectural design through ablation experiments on the three-tier memory model. Table~\ref{tab:ablation} presents results for configurations that selectively disable memory tiers, along with a single-tier baseline that combines all memory into one shared space. Removing ephemeral scratch increases memory by 1.5$\times$ and reduces accuracy by 2.9 points, as persisted intermediates degrade reasoning quality \cite{liu2024lost}. Disabling structured state yields the largest drop ($-$10.2 points), confirming its critical role in coherent multi-step reasoning. The external connector tier contributes 7.3 points by preventing context pollution from external data. The single-tier baseline degrades to 52.9\% accuracy with 2.0$\times$ higher memory, demonstrating that memory segregation is fundamental to GraphBit's effectiveness. Cross-platform results are in Appendix~\ref{sec:appendix_crossplatform}.

\begin{table}[t!]
\centering
\small
\resizebox{\columnwidth}{!}{%
\begin{tabular}{@{}lccc@{}}
\toprule
\textbf{Configuration} & \textbf{Acc. (\%)} & \textbf{Mem. (MB)} & \textbf{$\Delta$Acc.} \\
\midrule
Full GraphBit & \textbf{67.6} & \textbf{126.1} & --- \\
\quad w/o ephemeral scratch & 64.7 & 189.2 & $-$2.9 \\
\quad w/o structured state & 57.4 & 138.7 & $-$10.2 \\
\quad w/o external connectors & 60.3 & 130.4 & $-$7.3 \\
Single-tier baseline & 52.9 & 247.8 & $-$14.7 \\
\bottomrule
\end{tabular}}
\caption{Ablation study on the three-tier memory architecture. Each row disables one tier while retaining the others; the single-tier baseline combines all memory into one shared space. $\Delta$Acc.: accuracy change relative to full configuration.}
\label{tab:ablation}
\vspace{-1.5em}
\end{table}

\section{Discussion}
Our evaluation reveals that framework-induced hallucination, rather than LLM reasoning quality, is the dominant failure mode for prompt-orchestrated systems, with rates reaching 69.0\% on web-enabled tasks for LangGraph. GraphBit's 0\% hallucination rate combined with the highest accuracy challenges the assumption that LLM-based orchestration is necessary for flexible agent systems. Orchestration architecture matters most when tool routing is involved: frameworks perform comparably on zero-tool tasks but diverge substantially with tools. GraphBit achieves both the lowest latency and highest accuracy, with sub-linear overhead scaling from 6.0\,ms to 13.4\,ms, while the three-tier memory reduces token consumption (1,916 tokens/task vs.\ 6,276 for Pydantic~AI) and prevents reasoning degradation. Error analysis (Appendix~\ref{sec:appendix_error}) confirms zero orchestration-induced GraphBit failures, while 69.0\% of LangGraph web-task failures stem from hallucinated routing.

\section{Concluding Remarks}
GraphBit demonstrates that deterministic workflow orchestration eliminates reliability failures while enabling superior reasoning (67.6\% accuracy, 0\% hallucination, 11.9\,ms latency). By decoupling orchestration from the LLM and enforcing structured DAG execution, GraphBit is particularly suited for regulated settings requiring auditability and reproducibility. Despite these promising results, several limitations remain: GraphBit requires explicit DAG specification, our evaluation covers a single benchmark with limited Level~3 tasks, and identical LLM configurations may not reflect framework-specific tuning. Future work will explore hybrid deterministic LLM routing and broader benchmarks.

\bibliography{custom}

@article{yu2025dyntaskmas,
  title={{DynTaskMAS: A Dynamic Task Graph-driven Framework for Asynchronous and Parallel LLM-based Multi-Agent Systems}},
  author={Junwei Yu and Yepeng Ding and Hiroyuki Sato},
  year={2025},
  arxiv={2503.07675}
}

@misc{ke2025surveyfrontiersllmreasoning,
      title={A Survey of Frontiers in LLM Reasoning: Inference Scaling, Learning to Reason, and Agentic Systems}, 
      author={Zixuan Ke and Fangkai Jiao and Yifei Ming and Xuan-Phi Nguyen and Austin Xu and Do Xuan Long and Minzhi Li and Chengwei Qin and Peifeng Wang and Silvio Savarese and Caiming Xiong and Shafiq Joty},
      year={2025},
      eprint={2504.09037},
      archivePrefix={arXiv},
      primaryClass={cs.AI},
      url={https://arxiv.org/abs/2504.09037}, 
}

@inproceedings{hong2023metagpt,
  title={MetaGPT: Meta programming for a multi-agent collaborative framework},
  author={Hong, Sirui and Zhuge, Mingchen and Chen, Jonathan and Zheng, Xiawu and Cheng, Yuheng and Wang, Jinlin and Zhang, Ceyao and Wang, Zili and Yau, Steven Ka Shing and Lin, Zijuan and others},
  booktitle={The Twelfth International Conference on Learning Representations},
  year={2023}
}

@article{guo2024large,
  title={Large language model based multi-agents: A survey of progress and challenges},
  author={Guo, Taicheng and Chen, Xiuying and Wang, Yaqi and Chang, Ruidi and Pei, Shichao and Chawla, Nitesh V and Wiest, Olaf and Zhang, Xiangliang},
  journal={arXiv preprint arXiv:2402.01680},
  year={2024}
}

@inproceedings{mialon2023gaia,
  title={Gaia: a benchmark for general ai assistants},
  author={Mialon, Gr{\'e}goire and Fourrier, Cl{\'e}mentine and Wolf, Thomas and LeCun, Yann and Scialom, Thomas},
  booktitle={The Twelfth International Conference on Learning Representations},
  year={2023}
}

@article{boiko2023autonomous,
  title={Autonomous chemical research with large language models},
  author={Boiko, Daniil A and MacKnight, Robert and Kline, Ben and Gomes, Gabe},
  journal={Nature},
  volume={624},
  number={7992},
  pages={570--578},
  year={2023},
  publisher={Nature Publishing Group UK London}
}

@inproceedings{wu2024autogen,
  title={Autogen: Enabling next-gen LLM applications via multi-agent conversations},
  author={Wu, Qingyun and Bansal, Gagan and Zhang, Jieyu and Wu, Yiran and Li, Beibin and Zhu, Erkang and Jiang, Li and Zhang, Xiaoyun and Zhang, Shaokun and Liu, Jiale and others},
  booktitle={First Conference on Language Modeling},
  year={2024}
}

@article{annam2025langchain,
  title={Langchain: Simplifying development with language models},
  author={Annam, Sangeetha and Saxena, Merry and Kaushik, Ujjwal and Mittal, Shikha},
  journal={Textual Intelligence: Large Language Models and Their Real-World Applications},
  pages={287--304},
  year={2025},
  publisher={Wiley Online Library}
}

@inproceedings{qian2024chatdev,
  title={Chatdev: Communicative agents for software development},
  author={Qian, Chen and Liu, Wei and Liu, Hongzhang and Chen, Nuo and Dang, Yufan and Li, Jiahao and Yang, Cheng and Chen, Weize and Su, Yusheng and Cong, Xin and others},
  booktitle={Proceedings of the 62nd Annual Meeting of the Association for Computational Linguistics (Volume 1: Long Papers)},
  pages={15174--15186},
  year={2024}
}

@article{wang2024agent,
  title={Agent ai with langgraph: A modular framework for enhancing machine translation using large language models},
  author={Wang, Jialin and Duan, Zhihua},
  journal={arXiv preprint arXiv:2412.03801},
  year={2024}
}

@software{Liu_LlamaIndex_2022,
author = {Liu, Jerry},
doi = {10.5281/zenodo.1234},
month = {11},
title = {{LlamaIndex}},
url = {https://github.com/jerryjliu/llama_index},
year = {2022}
}

@inproceedings{yao2022react,
  title={React: Synergizing reasoning and acting in language models},
  author={Yao, Shunyu and Zhao, Jeffrey and Yu, Dian and Du, Nan and Shafran, Izhak and Narasimhan, Karthik R and Cao, Yuan},
  booktitle={The eleventh international conference on learning representations},
  year={2022}
}

@article{yao2023tree,
  title={Tree of thoughts: Deliberate problem solving with large language models},
  author={Yao, Shunyu and Yu, Dian and Zhao, Jeffrey and Shafran, Izhak and Griffiths, Tom and Cao, Yuan and Narasimhan, Karthik},
  journal={Advances in neural information processing systems},
  volume={36},
  pages={11809--11822},
  year={2023}
}

@article{shinn2023reflexion,
  title={Reflexion: Language agents with verbal reinforcement learning},
  author={Shinn, Noah and Cassano, Federico and Gopinath, Ashwin and Narasimhan, Karthik and Yao, Shunyu},
  journal={Advances in Neural Information Processing Systems},
  volume={36},
  pages={8634--8652},
  year={2023}
}

@article{schick2023toolformer,
  title={Toolformer: Language models can teach themselves to use tools},
  author={Schick, Timo and Dwivedi-Yu, Jane and Dess{\`\i}, Roberto and Raileanu, Roberta and Lomeli, Maria and Hambro, Eric and Zettlemoyer, Luke and Cancedda, Nicola and Scialom, Thomas},
  journal={Advances in Neural Information Processing Systems},
  volume={36},
  pages={68539--68551},
  year={2023}
}

@article{qin2024tool,
  title={Tool learning with foundation models},
  author={Qin, Yujia and Hu, Shengding and Lin, Yankai and Chen, Weize and Ding, Ning and Cui, Ganqu and Zeng, Zheni and Zhou, Xuanhe and Huang, Yufei and Xiao, Chaojun and others},
  journal={ACM Computing Surveys},
  volume={57},
  number={4},
  pages={1--40},
  year={2024},
  publisher={ACM New York, NY}
}

@incollection{haines2022workflow,
  title={Workflow orchestration with apache airflow},
  author={Haines, Scott},
  booktitle={Modern Data Engineering with Apache Spark: A Hands-On Guide for Building Mission-Critical Streaming Applications},
  pages={255--295},
  year={2022},
  publisher={Springer}
}

@article{khattab2023dspy,
  title={Dspy: Compiling declarative language model calls into self-improving pipelines},
  author={Khattab, Omar and Singhvi, Arnav and Maheshwari, Paridhi and Zhang, Zhiyuan and Santhanam, Keshav and Vardhamanan, Sri and Haq, Saiful and Sharma, Ashutosh and Joshi, Thomas T and Moazam, Hanna and others},
  journal={arXiv preprint arXiv:2310.03714},
  year={2023}
}

@article{beurer2023prompting,
  title={Prompting is programming: A query language for large language models},
  author={Beurer-Kellner, Luca and Fischer, Marc and Vechev, Martin},
  journal={Proceedings of the ACM on Programming Languages},
  volume={7},
  number={PLDI},
  pages={1946--1969},
  year={2023},
  publisher={ACM New York, NY, USA}
}

@incollection{narayanan2024orchestrating,
  title={Orchestrating Data Engineering Pipelines using Prefect},
  author={Narayanan, Pavan Kumar},
  booktitle={Data Engineering for Machine Learning Pipelines: From Python Libraries to ML Pipelines and Cloud Platforms},
  pages={415--449},
  year={2024},
  publisher={Springer}
}

@article{liu2024lost,
  title={Lost in the middle: How language models use long contexts},
  author={Liu, Nelson F and Lin, Kevin and Hewitt, John and Paranjape, Ashwin and Bevilacqua, Michele and Petroni, Fabio and Liang, Percy},
  journal={Transactions of the association for computational linguistics},
  volume={12},
  pages={157--173},
  year={2024}
}

@article{wei2022chain,
  title={Chain-of-thought prompting elicits reasoning in large language models},
  author={Wei, Jason and Wang, Xuezhi and Schuurmans, Dale and Bosma, Maarten and Xia, Fei and Chi, Ed and Le, Quoc V and Zhou, Denny and others},
  journal={Advances in neural information processing systems},
  volume={35},
  pages={24824--24837},
  year={2022}
}

@article{duan2024exploration,
  title={Exploration of llm multi-agent application implementation based on langgraph+ crewai},
  author={Duan, Zhihua and Wang, Jialin},
  journal={arXiv preprint arXiv:2411.18241},
  year={2024}
}

@article{zhang2025survey,
  title={A survey on the memory mechanism of large language model-based agents},
  author={Zhang, Zeyu and Dai, Quanyu and Bo, Xiaohe and Ma, Chen and Li, Rui and Chen, Xu and Zhu, Jieming and Dong, Zhenhua and Wen, Ji-Rong},
  journal={ACM Transactions on Information Systems},
  volume={43},
  number={6},
  pages={1--47},
  year={2025},
  publisher={ACM New York, NY}
}

@inproceedings{kim2024llm,
  title={An llm compiler for parallel function calling},
  author={Kim, Sehoon and Moon, Suhong and Tabrizi, Ryan and Lee, Nicholas and Mahoney, Michael W and Keutzer, Kurt and Gholami, Amir},
  booktitle={Forty-first International Conference on Machine Learning},
  year={2024}
}

@article{patil2024gorilla,
  title={Gorilla: Large language model connected with massive apis},
  author={Patil, Shishir G and Zhang, Tianjun and Wang, Xin and Gonzalez, Joseph E},
  journal={Advances in Neural Information Processing Systems},
  volume={37},
  pages={126544--126565},
  year={2024}
}

@article{polyzotis2019data,
  title={Data validation for machine learning},
  author={Polyzotis, Neoklis and Zinkevich, Martin and Roy, Sudip and Breck, Eric and Whang, Steven},
  journal={Proceedings of machine learning and systems},
  volume={1},
  pages={334--347},
  year={2019}
}

@article{sculley2015hidden,
  title={Hidden technical debt in machine learning systems},
  author={Sculley, David and Holt, Gary and Golovin, Daniel and Davydov, Eugene and Phillips, Todd and Ebner, Dietmar and Chaudhary, Vinay and Young, Michael and Crespo, Jean-Francois and Dennison, Dan},
  journal={Advances in neural information processing systems},
  volume={28},
  year={2015}
}

@article{cemri2025multi,
  title={Why do multi-agent llm systems fail?},
  author={Cemri, Mert and Pan, Melissa Z and Yang, Shuyi and Agrawal, Lakshya A and Chopra, Bhavya and Tiwari, Rishabh and Keutzer, Kurt and Parameswaran, Aditya and Klein, Dan and Ramchandran, Kannan and others},
  journal={arXiv preprint arXiv:2503.13657},
  year={2025}
}

@article{qiu2025blueprint,
  title={Blueprint First, Model Second: A Framework for Deterministic LLM Workflow},
  author={Qiu, Libin and Ye, Yuhang and Gao, Zhirong and Zou, Xide and Chen, Junfu and Gui, Ziming and Huang, Weizhi and Xue, Xiaobo and Qiu, Wenkai and Zhao, Kun},
  journal={arXiv preprint arXiv:2508.02721},
  year={2025}
}

@article{masterman2024landscape,
  title={The landscape of emerging ai agent architectures for reasoning, planning, and tool calling: A survey},
  author={Masterman, Tula and Besen, Sandi and Sawtell, Mason and Chao, Alex},
  journal={arXiv preprint arXiv:2404.11584},
  year={2024}
}

@article{qin2023toolllm,
  title={Toolllm: Facilitating large language models to master 16000+ real-world apis},
  author={Qin, Yujia and Liang, Shihao and Ye, Yining and Zhu, Kunlun and Yan, Lan and Lu, Yaxi and Lin, Yankai and Cong, Xin and Tang, Xiangru and Qian, Bill and others},
  journal={arXiv preprint arXiv:2307.16789},
  year={2023}
}

@inproceedings{dennis2005first,
  title={First version of a data flow procedure language},
  author={Dennis, Jack B},
  booktitle={Programming Symposium: Proceedings, Colloque sur la Programmation Paris, April 9--11, 1974},
  pages={362--376},
  year={2005},
  organization={Springer}
}

@article{thost2021directed,
  title={Directed acyclic graph neural networks},
  author={Thost, Veronika and Chen, Jie},
  journal={arXiv preprint arXiv:2101.07965},
  year={2021}
}

@article{qiao2024benchmarking,
  title={Benchmarking agentic workflow generation},
  author={Qiao, Shuofei and Fang, Runnan and Qiu, Zhisong and Wang, Xiaobin and Zhang, Ningyu and Jiang, Yong and Xie, Pengjun and Huang, Fei and Chen, Huajun},
  journal={arXiv preprint arXiv:2410.07869},
  year={2024}
}

@inproceedings{ranaldi2024tree,
  title={A tree-of-thoughts to broaden multi-step reasoning across languages},
  author={Ranaldi, Leonardo and Pucci, Giulia and Ranaldi, Federico and Ruzzetti, Elena Sofia and Zanzotto, Fabio Massimo},
  booktitle={Findings of the Association for Computational Linguistics: NAACL 2024},
  pages={1229--1241},
  year={2024}
}

@article{gu2025large,
  title={Large language models for constructing and optimizing machine learning workflows: A survey},
  author={Gu, Yang and You, Hengyu and Cao, Jian and Yu, Muran and Fan, Haoran and Qian, Shiyou},
  journal={ACM Transactions on Software Engineering and Methodology},
  year={2025},
  publisher={ACM New York, NY}
}

@article{tran2025multi,
  title={Multi-agent collaboration mechanisms: A survey of llms},
  author={Tran, Khanh-Tung and Dao, Dung and Nguyen, Minh-Duong and Pham, Quoc-Viet and O'Sullivan, Barry and Nguyen, Hoang D},
  journal={arXiv preprint arXiv:2501.06322},
  year={2025}
}

@article{barua2024exploring,
  title={Exploring autonomous agents through the lens of large language models: A review},
  author={Barua, Saikat},
  journal={arXiv preprint arXiv:2404.04442},
  year={2024}
}

@article{bugden2022rust,
  title={Rust: The programming language for safety and performance},
  author={Bugden, William and Alahmar, Ayman},
  journal={arXiv preprint arXiv:2206.05503},
  year={2022}
}

@article{habib2019type,
  title={Type safety with JSON subschema},
  author={Habib, Andrew and Shinnar, Avraham and Hirzel, Martin and Pradel, Michael},
  journal={arXiv preprint arXiv:1911.12651},
  year={2019}
}

\appendix

\section{Implementation Details}
\label{sec:appendix_implementation}

GraphBit's execution engine is implemented in approximately 8,000 lines of Rust code, with an additional 2,000 lines of Python bindings generated via PyO3. The core components include a workflow graph representation using the \texttt{petgraph} library, a thread pool executor based on \texttt{tokio}, and a state management system using \texttt{serde} for serialization. Agent nodes interface with LLM providers through a unified abstraction layer supporting OpenAI, Anthropic, and local model deployments. Tool nodes implement an asynchronous execution interface with configurable timeout and retry policies. Control nodes evaluate predicates using a simple expression language supporting boolean operations over typed state variables.

\section{Benchmark Configuration}
\label{sec:appendix_benchmark}
The primary benchmark experiments were conducted across multiple cloud and bare-metal systems spanning diverse processor architectures: System~A (Intel Xeon Skylake, 2 vCPUs, 2 GiB RAM, Ubuntu), System~B (AMD EPYC 7571, 2 vCPUs, 2 GiB RAM, Ubuntu), System~C (Intel Xeon Skylake, 2 vCPUs, 4 GiB RAM, Windows), System~D (AMD EPYC 7571, 2 vCPUs, 4 GiB RAM, Windows), and System~E (Apple M1, 8 vCPUs, 16 GiB RAM, macOS). Each framework received equivalent computational resources and API rate limits to ensure fair comparison. All frameworks used the same proprietary LLM with identical temperature (1.0) and max\_tokens (2,000) parameters. To validate cross-platform consistency, we additionally conducted ablation experiments across three distinct hardware configurations: Apple Mac M4 (ARM architecture, 16GB RAM, macOS Sequoia), Ubuntu 22.04 on Intel Xeon W-2255 (x86-64, 64GB RAM), and Windows 11 on Intel Core i9-13900K (x86-64, 32GB RAM).

We extend the standard GAIA evaluation protocol with six metrics. Accuracy measures the percentage of correctly completed tasks via dual verification: exact string matching against ground truth and independent LLM-based evaluation using GPT-5.2-chat as an evaluator. Hallucination rate measures the percentage of failed executions due to framework-induced errors (routing failures, infinite loops, runtime crashes). Processing time reports the mean framework overhead in milliseconds, excluding LLM API latency, measured via the \texttt{psutil} library. CPU utilization measures average processor usage during execution. Peak memory reports the highest memory consumption in megabytes. Throughput is computed as $60000 / \text{processing\_time\_ms}$ operations per minute.

For each framework, we implemented equivalent agent configurations using framework-idiomatic patterns across three workflow types. Zero-tool workflows use direct LLM prompting. Document-augmented workflows employ framework-native tool-calling agents with 12 custom tools including PDF reading (PyPDF2), Excel processing (pandas), image analysis (Donut vision encoder-decoder), audio transcription (Whisper), and code execution. Web-enabled workflows use DuckDuckGo search with BeautifulSoup parsing. Agent execution was limited to max\_iterations=3 across all frameworks for fair comparison.

\paragraph{Cross-Platform Ablation Setup}

The ablation experiments were conducted across three hardware configurations to validate cross-platform consistency:

\begin{itemize}
\item \textbf{Mac M4 (ARM)}: Apple Mac Mini M4 with 16GB unified memory, macOS Sequoia 15.1, Rust 1.75.0, Python 3.11.7
\item \textbf{Ubuntu Intel (x86-64)}: Intel Xeon W-2255 with 64GB DDR4 RAM, Ubuntu 22.04.3 LTS, Rust 1.75.0, Python 3.11.7
\item \textbf{Windows Intel (x86-64)}: Intel Core i9-13900K with 32GB DDR5 RAM, Windows 11 Pro 23H2, Rust 1.75.0, Python 3.11.7
\end{itemize}

Each platform ran identical GraphBit configurations compiled from the same source revision. Python dependencies were pinned to identical versions across all environments using Poetry lock files. LLM API calls were routed through a centralized proxy to ensure consistent network latency measurements.

\paragraph{Workflow Configuration}

GraphBit workflows were constructed with the following structure: an initial planning agent determines the task decomposition, followed by parallel execution of subtask agents, and a final synthesis agent aggregates results. Conditional branches handle error recovery and alternative approaches when initial attempts fail. Baseline frameworks were configured according to their documentation best practices. LangChain used sequential chains with ReAct agents. LangGraph used state graphs with conditional edges. CrewAI used hierarchical crew structures. AutoGen used two-agent conversation patterns. Pydantic AI used typed agent runs. LlamaIndex used query engine agents with tool augmentation.

\section{Performance by Difficulty Level}
\label{sec:appendix_difficulty}

Table~\ref{tab:gaia_levels} presents accuracy by GAIA difficulty level for all seven frameworks. GraphBit achieves the highest accuracy on Level~1 (79.3\%) and Level~2 (63.9\%) tasks. LangGraph drops sharply from 48.3\% (Level~1) to 27.8\% (Level~2), and both LangGraph and AutoGen reach 0\% on Level~3, suggesting that prompt-orchestrated routing compounds errors as task complexity increases.

\begin{table}[ht!]
\centering
\small
\resizebox{0.8\columnwidth}{!}{%
\begin{tabular}{@{}lccc@{}}
\toprule
\textbf{Framework} & \textbf{Level 1} & \textbf{Level 2} & \textbf{Level 3} \\
 & (n=29) & (n=36) & (n=3) \\
\midrule
LangChain & 41.4 & 33.3 & 33.3 \\
LangGraph & 48.3 & 27.8 & 0.0 \\
CrewAI & 51.7 & 63.9 & 66.7 \\
AutoGen & 58.6 & 38.9 & 0.0 \\
Pydantic AI & 62.1 & 63.9 & 66.7 \\
LlamaIndex & 62.1 & 44.4 & 66.7 \\
\midrule
\textbf{GraphBit} & \textbf{79.3} & \textbf{63.9} & \textbf{66.7} \\
\bottomrule
\end{tabular}}
\caption{Accuracy (\%) by GAIA difficulty level. Level~1: simple single-step tasks; Level~2: moderate multi-step reasoning; Level~3: complex planning with extensive tool use. Level~3 contains only 3 tasks; results should be interpreted with caution.}
\label{tab:gaia_levels}
\end{table}

\section{Workflow Reliability Details}
\label{sec:appendix_reliability}

Table~\ref{tab:reliability} presents the complete hallucination rate breakdown by task type for all seven frameworks.

\begin{table}[ht!]
\centering
\small
\resizebox{0.9\columnwidth}{!}{%
\begin{tabular}{@{}lcccc@{}}
\toprule
\textbf{Framework} & \textbf{No-Tool} & \textbf{Local} & \textbf{Web} & \textbf{Overall} \\
\midrule
LangChain & 0.0 & 15.8 & 59.5 & 41.2 \\
LangGraph & 0.0 & 15.8 & 69.0 & 47.1 \\
CrewAI & 0.0 & 0.0 & 16.7 & 14.3 \\
AutoGen & 0.0 & 42.1 & 35.7 & 33.8 \\
Pydantic AI & 0.0 & 0.0 & 0.0 & 0.0 \\
LlamaIndex & 0.0 & 0.0 & 0.0 & 0.0 \\
\midrule
\textbf{GraphBit} & \textbf{0.0} & \textbf{0.0} & \textbf{0.0} & \textbf{0.0} \\
\bottomrule
\end{tabular}}
\caption{Hallucination rate (\%) by task type. Zero-tool tasks produce no hallucinations across all frameworks. Hallucination rates escalate with task complexity, particularly for prompt-orchestrated frameworks on web-enabled tasks.}
\label{tab:reliability}
\end{table}

\section{Additional Results}
\label{sec:appendix_results}

\subsection{Complete Computational Efficiency by Task Type}

Table~\ref{tab:full_efficiency} presents the complete computational efficiency metrics for all seven frameworks across the three task types, measured on correctly completed tasks. GraphBit achieves the lowest processing latency across all task types. The efficiency advantage is most pronounced on no-tool tasks (6.0\,ms, 3.9$\times$ faster than AutoGen) and document-augmented tasks (10.8\,ms, 14.7$\times$ faster than AutoGen at 159.1\,ms). AutoGen's exceptionally high processing time on document-augmented tasks reflects its conversation-based orchestration pattern, which requires multiple LLM round-trips for tool coordination \cite{kim2024llm}.

\begin{table*}[ht!]
\centering
\small
\resizebox{0.8\textwidth}{!}{%
\begin{tabular}{@{}llcccccc@{}}
\toprule
\textbf{Task Type} & \textbf{Framework} & \textbf{Proc. (ms)} & \textbf{CPU (\%)} & \textbf{Mem. (MB)} & \textbf{Thpt. (ops/m)} & \textbf{$n$} \\
\midrule
\multirow{7}{*}{No-Tool} & GraphBit & \textbf{6.0} & \textbf{15.2} & \textbf{34.9} & \textbf{10,000} & 5 \\
 & LangChain & 7.6 & 18.2 & 86.7 & 7,887 & 4 \\
 & CrewAI & 9.4 & 17.3 & 58.1 & 6,353 & 3 \\
 & LlamaIndex & 11.2 & 20.3 & 114.6 & 5,361 & 5 \\
 & Pydantic AI & 12.0 & 19.0 & 186.2 & 5,002 & 2 \\
 & LangGraph & 16.9 & 18.3 & 106.2 & 3,550 & 4 \\
 & AutoGen & 23.4 & 19.3 & 47.0 & 2,562 & 4 \\
\midrule
\multirow{7}{*}{Local Tool} & GraphBit & \textbf{10.8} & \textbf{24.7} & \textbf{105.5} & \textbf{5,543} & 11 \\
 & LlamaIndex & 15.3 & 29.6 & 160.9 & 3,926 & 13 \\
 & CrewAI & 16.1 & 26.4 & 151.5 & 3,728 & 8 \\
 & Pydantic AI & 16.3 & 26.5 & 135.5 & 3,691 & 12 \\
 & LangChain & 41.8 & 30.1 & 200.5 & 1,435 & 10 \\
 & LangGraph & 46.3 & 33.7 & 199.6 & 1,295 & 11 \\
 & AutoGen & 159.1 & 32.7 & 171.3 & 377 & 7 \\
\midrule
\multirow{7}{*}{Web Search} & GraphBit & \textbf{13.4} & \textbf{20.5} & \textbf{150.5} & \textbf{4,466} & 32 \\
 & LlamaIndex & 15.5 & 24.0 & 175.9 & 3,868 & 18 \\
 & Pydantic AI & 20.4 & 24.1 & 177.3 & 2,948 & 29 \\
 & LangGraph & 27.3 & 24.0 & 228.8 & 2,201 & 9 \\
 & CrewAI & 34.6 & 32.9 & 226.2 & 1,736 & 29 \\
 & AutoGen & 37.5 & 25.7 & 359.7 & 1,599 & 20 \\
 & LangChain & 38.3 & 23.4 & 274.4 & 1,567 & 11 \\
\bottomrule
\end{tabular}}
\caption{Complete computational efficiency metrics by task type (measured on correctly completed tasks only). $n$: number of correctly completed tasks. Frameworks sorted by processing time within each task type.}
\label{tab:full_efficiency}
\end{table*}

\subsection{Cross-Platform Consistency}
\label{sec:appendix_crossplatform}

Table~\ref{tab:ablation_platforms} presents GraphBit's performance across three hardware platforms to validate cross-platform consistency of the Rust-based execution engine.

\begin{table}[ht!]
\centering
\small
\resizebox{0.95\columnwidth}{!}{%
\begin{tabular}{@{}lccc@{}}
\toprule
\textbf{Platform} & \textbf{Acc. (\%)} & \textbf{Mem. (MB)} & \textbf{Proc. (ms)} \\
\midrule
Mac M4 (ARM) & 67.4 & 118.3 & 11.7 \\
Ubuntu Intel (x86) & 67.6 & 126.1 & 11.9 \\
Windows Intel (x86) & 67.9 & 132.8 & 12.4 \\
\midrule
\textit{Std. Dev.} & \textit{0.3} & \textit{7.3} & \textit{0.4} \\
\bottomrule
\end{tabular}}
\caption{Cross-platform consistency of GraphBit across Mac M4 (ARM), Ubuntu Intel, and Windows Intel. Accuracy variations remain within 0.5 percentage points.}
\label{tab:ablation_platforms}
\end{table}

Accuracy variations remain within 0.5 percentage points across platforms, confirming that our architectural advantages are not artifacts of a specific runtime environment. The Mac M4's unified memory architecture yields 6\% lower memory consumption than Ubuntu Intel, while Windows exhibits marginally higher overhead due to additional runtime dependencies.

\subsection{Token Efficiency Analysis}

Table~\ref{tab:token_efficiency} reports token consumption patterns across frameworks. Token efficiency directly impacts API cost and latency in production deployments.

\begin{table}[ht!]
\centering
\small
\resizebox{0.9\columnwidth}{!}{%
\begin{tabular}{@{}lcccc@{}}
\toprule
\textbf{Framework} & \textbf{Prompt} & \textbf{Compl.} & \textbf{Total} & \textbf{TPS} \\
\midrule
GraphBit & 1,735 & 181 & 1,916 & 12.9 \\
Pydantic AI & 6,004 & 273 & 6,276 & 17.0 \\
CrewAI & 11,306 & 2,332 & 13,638 & 33.2 \\
LangChain & 253 & 176 & 180 & 35.9 \\
LangGraph & 237 & 111 & 174 & 31.6 \\
\bottomrule
\end{tabular}}
\caption{Mean token consumption per task. Prompt: mean prompt tokens; Compl.: mean completion tokens; Total: mean total tokens; TPS: tokens per second. Frameworks with incomplete token reporting are omitted.}
\label{tab:token_efficiency}
\end{table}

GraphBit consumes 1,916 mean total tokens per task, 3.3$\times$ fewer than Pydantic~AI (6,276) and 7.1$\times$ fewer than CrewAI (13,638). This efficiency stems from GraphBit's structured state management, which avoids the context accumulation patterns of conversation-based frameworks. CrewAI's high token consumption reflects its verbose agent interaction protocol, where backstory and role definitions are repeated across multiple agent turns. LangChain and LangGraph report lower total tokens but this reflects their use of sequential chains that make fewer but larger LLM calls.

\subsection{Execution Time Distribution}

Table~\ref{tab:latency_dist} characterizes the execution time distribution across frameworks, providing insight into predictability and tail latency behavior.

\begin{table}[ht!]
\centering
\small
\resizebox{0.98\columnwidth}{!}{%
\begin{tabular}{@{}lccccc@{}}
\toprule
\textbf{Framework} & \textbf{Mean} & \textbf{Med.} & \textbf{P95} & \textbf{Std.} & \textbf{Max} \\
 & (s) & (s) & (s) & (s) & (s) \\
\midrule
LlamaIndex & 27.9 & 24.5 & 70.5 & 23.4 & 95.1 \\
LangGraph & 27.4 & 17.3 & 73.9 & 25.2 & 84.9 \\
GraphBit & 35.3 & 28.8 & 115.7 & 34.8 & 140.0 \\
Pydantic AI & 38.7 & 29.5 & 101.8 & 39.7 & 213.8 \\
LangChain & 45.1 & 37.5 & 119.8 & 42.4 & 168.1 \\
AutoGen & 54.9 & 53.1 & 142.6 & 49.2 & 175.0 \\
CrewAI & 75.8 & 50.7 & 217.5 & 80.8 & 347.0 \\
\bottomrule
\end{tabular}}
\caption{End-to-end execution time distribution across all 68 tasks (including LLM API latency). Med.: median; P95: 95th percentile; Std.: standard deviation.}
\label{tab:latency_dist}
\end{table}

End-to-end execution time is dominated by LLM API latency rather than framework processing overhead. LlamaIndex and LangGraph achieve the lowest mean execution times (27.9\,s and 27.4\,s), though their lower times partly reflect fewer successful tool invocations on failed tasks. CrewAI exhibits the highest variance (80.8\,s standard deviation) and worst tail latency (347.0\,s maximum), attributable to its multi-agent conversation protocol which can trigger extended deliberation chains. GraphBit's P95 latency of 115.7\,s is higher than LlamaIndex (70.5\,s), reflecting GraphBit's more thorough task processing enabled by successful tool chains on complex tasks.

\subsection{Error Analysis}
\label{sec:appendix_error}

Of the 22 tasks (32.4\%) where GraphBit failed to produce correct answers, manual analysis reveals the following distribution: 50\% involved factual errors in LLM reasoning where the model generated incorrect intermediate conclusions; 30\% involved misinterpretation of task requirements, particularly ambiguous natural language specifications; 15\% involved tool execution failures such as web search returning no relevant results; and 5\% involved output formatting errors where correct reasoning produced incorrectly formatted final answers. Critically, zero failures were attributable to orchestration errors, confirming that GraphBit's execution engine operates correctly across all evaluated scenarios. In contrast, baseline framework failures exhibit a bimodal distribution between orchestration errors and reasoning errors. For LangGraph, 69.0\% of web-search task failures are orchestration-induced (hallucinated tool routing) \cite{patil2024gorilla}, while only 31.0\% are reasoning errors. This finding suggests that a substantial fraction of baseline failures could be eliminated through deterministic orchestration alone, without requiring improvements to the underlying LLM.

\subsection{Framework Component Overhead}

Table~\ref{tab:setup_overhead} reports framework initialization overhead, which impacts cold-start latency in serverless deployments.CrewAI incurs the highest import overhead (5,700\,ms) due to its extensive dependency chain, while Pydantic~AI achieves the lowest (2,100\,ms). AutoGen's setup time (23.6\,ms) is substantially higher than other frameworks due to its agent initialization protocol. GraphBit's combined initialization of 2,400\,ms is competitive despite the overhead of loading Rust bindings via PyO3.

\begin{table}[t!]
\centering
\small
\resizebox{0.8\columnwidth}{!}{%
\begin{tabular}{@{}lcc@{}}
\toprule
\textbf{Framework} & \textbf{Import (ms)} & \textbf{Setup (ms)} \\
\midrule
GraphBit & 2,400 & 0.1 \\
LangChain & 3,200 & 0.9 \\
LangGraph & 2,600 & 0.9 \\
LlamaIndex & 2,900 & 0.2 \\
Pydantic AI & 2,100 & 1.2 \\
AutoGen & 3,700 & 23.6 \\
CrewAI & 5,700 & 0.1 \\
\bottomrule
\end{tabular}}
\caption{Framework initialization overhead. Import: Python module import time; Setup: framework-specific initialization after import.}
\label{tab:setup_overhead}
\end{table}

\end{document}